# Towards Automatic Digital Documentation and Progress Reporting of Mechanical Construction Pipes using Smartphones

Reza Maalek[1, *], Derek D. Lichti[2] and Shahrokh Maalek[3]

[1] Endowed Professor and Chair of Digital Engineering and Construction, Institute for Technology and Management in Construction (TMB), Karlsruhe Institute of Technology (KIT). Address: Am Fasanengarten, Karlsruhe, Germany, 76131. Email: reza.maalek@kit.edu (corresponding author)

[2] Professor of Geomatics Engineering, Department of Geomatics Engineering of the Schulich School of Engineering, University of Calgary. Address: 2500 University Dr. NW, Calgary, Canada, T2N 1N4. Email: ddlichti@ucalgary.ca

[3] Chief Technology Officer (CTO), Digital Information in Construction Engineering (DICE) Technologies. Address: 150 9 Ave SW, Calgary, Canada, T2P 3H9. Email: shahrokh.maalek@dicetechnologies.ca

* Indicates corresponding author

**Abstract:** This manuscript presents a new framework towards automated digital documentation and progress reporting of mechanical pipes in building construction projects, using smartphones. New methods were proposed to optimize video frame rate to achieve a desired image overlap; define metric scale for 3D reconstruction; extract pipes from point clouds; and classify pipes according to their planned bill of quantity radii. The effectiveness of the proposed methods in both laboratory (six pipes) and construction site (58 pipes) conditions was evaluated. It was observed that the proposed metric scale definition achieved sub-millimeter pipe radius estimation accuracy. Both laboratory and field experiments revealed that increasing the defined image overlap improved point cloud quality, pipe classification quality, and pipe radius/length estimation. Overall, it was found possible to achieve pipe classification F-measure, radius estimation accuracy, and length estimation percent error of 96.4%, 5.4mm, and 5.0%, respectively, on construction sites using at least 95% image overlap.

**Keywords:** 3D metric reconstruction, smartphone photogrammetry, point cloud pipe detection, mechanical progress reporting, mechanical pipes as-built digital documentation

1. **Introduction: reporting progress of mechanical pipes**

The construction industry best practices in reporting the progress of mechanical pipes involve the estimation of the length of installed pipe (of a specific diameter and type). This information is typically gathered manually by the trades person in charge of pipe installation and fed to a central system (either through a tablet or mobile phone) for progress reporting, and project monitoring and control. The manual estimation of the length of installed pipe, however, contains an inherent margin of error. For instance, especially in larger residential and commercial construction with multiple crews, the lengths estimated by different work crews are often reported differently, creating an inconsistency, particularly since an industry standard and procedure for reporting the length does not currently exist. Furthermore, the time spent estimating the lengths of the installed pipes is not commonly considered as tool-time [1], which might increase reporting errors due to the lack of incentive to perform thorough measurements. In residential and commercial building construction, the margin of error in manual reporting of the length of installed pipe is estimated around ±15-20% [2,3]. The existence of this margin of error not only promotes an erroneous mechanical progress reporting during construction, but it can also: (i) hinder the ability of the mechanical contractors to provide competitive bids on prospective projects of similar sizes; and (ii) possibly compel the contractor to order additional unnecessary pipes and materials, which contributes to construction waste. An automated and accurate solution to estimating the length of installed pipes can, hence, (i) improve tool-time and progress reporting accuracy; (ii) serve as a historical database for competitive bidding purposes; and (iii) potentially reduce mechanical construction waste.

Recent advancements in computer vision and photogrammetry enable automated digital documentation and acquisition of accurate 3-dimensional (3D) point clouds of construction projects using optical instruments such as laser scanners [4] and cameras [5]. The point clouds acquired from mechanical pipes must then be further analysed and transformed into semantic information -in our case, converted into pipe lengths with respect to their diameters- before they can be effectively utilized for progress reporting and quality control. Current solutions for detecting and classifying pipes and their prospective lengths from point clouds, however, possess the following limitations:

1- Manual detection of pipe lengths from point clouds can be time consuming and might introduce an additional error source due to subjectivity of the manual intervention [4].
2- Commercially available automated solutions for point cloud processing in construction, such as Verity [6], and 2D3D RMap [7], require 2D, 3D or 4D building information models (BIMs), registered with the coordinate system of the point cloud, which may not be readily available with the required detail.
3- Laser scanners, especially terrestrial laser scanners (TLS), can be expensive and often require skilled personnel for operation, and data analysis. Therefore, the associated front-end expenditure as well as operating costs might not be feasible for many mechanical sub-contractors with lower profit margins.
4- Photogrammetric point clouds generated from images require an accurate scale definition for metric reconstruction, which is predominantly performed manually [5].
5- Even though high resolution point clouds are ideal for accurate as-built modeling and object detection [8], they impose an additional obstacle related to storage and analysis of big data.

To this end, this manuscript aims to address the aforementioned limitations by:

1- Utilizing video recordings from smartphone cameras to generate point clouds, a readily available device, requiring minimum front-end investment.
2- Defining a systematic process to extract sufficiently many images from video recordings to improve 3D reconstruction resolution and accuracy, while preserving computational efficiency.
3- Introducing an automated procedure to define accurate metric scale for photogrammetric point clouds on construction projects.
4- Developing a new method for detection of cylinders from photogrammetric point clouds, irrespective of the details of the planned BIM; and
5- Assessing the effectiveness of the proposed methods under controlled laboratory conditions as well as real-world construction sites.

## 1.1. Structure of manuscript

The remainder of the manuscript is organized as follows:

- Section 2: Literature review provides a brief overview of sequential structure-from-motion (SfM) for 3D reconstruction followed by the current state of research for photogrammetry in construction.
- Section 3: Methodology describes the proposed methods and algorithms followed by the metrics used for validation of the methodology.
- Section 4: Experimental design explains the design and development of the laboratory as well as real-world experiments used to assess the effectiveness of the proposed methods.
- Section 5: Experimental results outlines the results of each experiments followed by discussions.
- Section 6: Conclusions summarizes the findings and discusses the avenues for future research.

## 2. Literature review

The review of previous work is divided into two subsections, namely, dense 3D reconstruction from images, and photogrammetry in construction, with particular emphasis on the generation of point clouds from images along with the application of images in progress reporting. For further information the reader is referred to [9,10] for 3D reconstruction technologies in construction, [11–13] for progress reporting, [14,15] for as-built model generation and [4,16–18] for point cloud processing. The two subsections are discussed in more details in the following.

### 2.1. Dense 3D reconstruction from images

SfM is the process of recovering the exterior orientation parameters (EOPs), i.e. position and orientation, as well as interior orientation parameters (IOPs), i.e. calibration parameters (in case of uncalibrated cameras) of overlapping image views, subjected to a rigid body motion (rotation and translation; [19]). The process requires sufficiently many point correspondences between two or more image views to provide a solution. For instance, given two image views with at least seven image point correspondences (in the uncalibrated camera case), it is possible to estimate the fundamental matrix [20] and consequentially recover the camera projective matrices up to a projective ambiguity[1] [21]. For calibrated cameras, five point correspondences between two images is sufficient to recover the relative orientation (rotation and translation [22]), and provide 3D reconstruction up to a similarity transform (i.e. an arbitrary scale [23]). The EOPs and IOPs of a given set of overlapping images can be formulated as either a global optimization problem[2] or by sequential reconstruction[3]. In practice, due to the flexibility and control inherent in sequential reconstruction and registration of new images, sequential SfM is expected to produce more reliable results than global [24]. Once the EOPs (and IOPs for uncalibrated cameras) are determined, dense pixel to pixel matching is performed to generate a dense 3D reconstruction of a scene. The process for sequential SfM along with dense 3D reconstruction is schematically presented in Figure 1 and can be summarized in the following stages:

1- Detect feature points in each images (Figure 1a; typically performed using scale-invariant feature transform (SIFT) [25]).
2- Match features between every two images (Figure 1b).
3- Select the best pair of images (i.e. images obtaining the highest score for some geometric selection criteria [24]), and estimate the relative orientation parameters and camera matrices between the two images using the matched feature points.
4- Determine the 3D coordinates of the corresponding points using triangulation.
5- Perform bundle adjustment to refine the camera parameters.
6- Add new images one by one and perform steps 3 through 5 until all images are sequentially registered.
7- The result of the aforementioned steps is a set of EOPs (and IOPs for uncalibrated cameras) that observe the coarse (sparse) 3D reconstructed points (tie points used for image registration) shown in Figure 1c.
8- Using the estimated camera parameters, dense pixel to pixel matching between every two images is carried out to generate a dense 3D reconstruction (Figure 1d).

---

[1] If the same camera is used between the views (i.e. same IOPs), which is the case in rigid body motion, the reconstruction is possible up to an affine transformation [23].
[2] i.e. all correspondences and constraints of all images are used together such as the projective factorization method of [68].
[3] Starting with a good initial pair of images and addition of images one by one for sequential reconstruction and registration.

SfM can provide 3D reconstruction up to an arbitrary scale factor only if an initial estimate of the IOPs along with the appropriate camera model are known a priori[4]. In [26], it was observed that SfM performed using initial IOPs obtained from laboratory pre-calibration, outperformed that obtained from exchangeable image file (Exif) for pipe reconstruction from smartphone cameras. Hence, in this manuscript, the smartphone cameras are pre-calibrated using the methods provided in [26,27].

Other than the importance of reliable IOPs, the number along with the geometry of matched feature points play crucial roles in recovering the camera parameters [28]. For instance, Zhang [20] showed that point correspondences spread throughout images, compared to isolated set of corresponding points, provides a more accurate estimation of the fundamental matrix between two images. Therefore, a random and sporadic set of images obtained from a scene cannot generally guarantee that: (i) sufficiently redundant set of matching points exist between every two images; and (ii) the geometry of the matched points is overall strong. The latter characteristics are less likely to transpire in videos due to the multiple frames that are captured per second. In fact, the advantage of videos compared to sporadic images was empirically corroborated in the work of Rebolj [29] when assessing the point cloud quality requirements for construction progress monitoring.

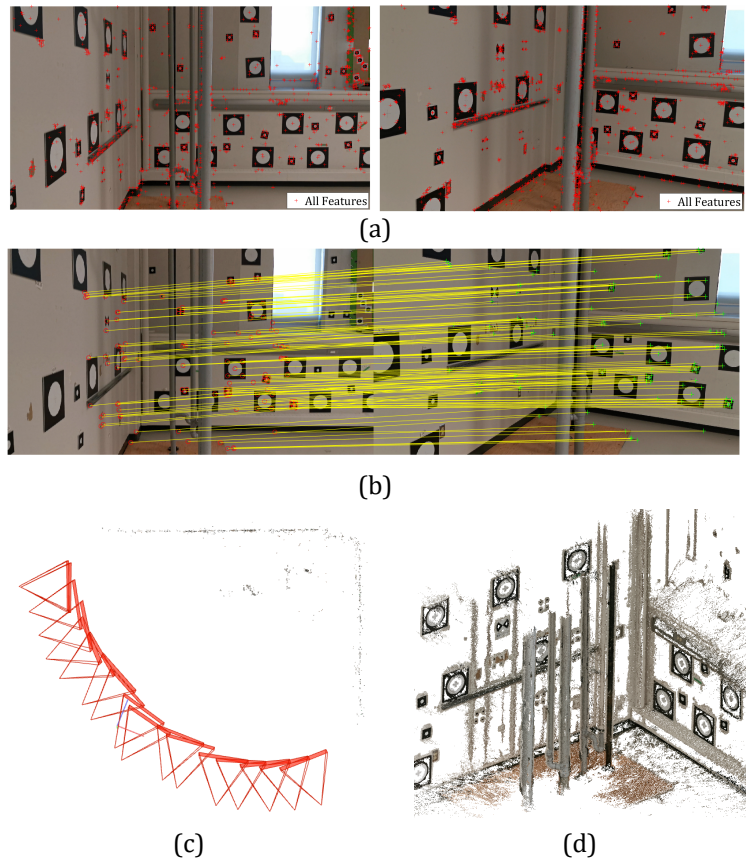

**Figure 1:** Process of dense 3D reconstruction from video sequences: a) detected features in each image; b) matched features between images; c) coarse reconstruction (tie points for image registration) and estimation of EOPs (and IOPs for uncalibrated cameras); d) dense 3D reconstruction

Recent smartphone cameras, such as the Huawei P30 used in this study, can acquire 4K video recordings with up to 60 frames per second (fps). 60 fps suggests that every second of the video can be decomposed into

---

[4] The IOPs and camera model are required in step 3 above to decompose the fundamental matrix into the essential matrix and recover the relative orientation parameters between cameras (see [23,24]).

60 4K images, which for normal walking speeds (around 1.4-1.5 m/s [30]) can likely provide a highly redundant network of point correspondences to ensure convergence. However, in practice, the 60 fps introduces additional problems with computation time for 3D reconstruction as well as storage. For instance, Golparvar-Fard [31] reported about 7 hours of processing time for dense reconstruction from 242 images, which is roughly 4 seconds of video recording decomposed at 60 fps. From the authors' recent experiences with COLMAP [24], a reliable open source SfM software used in this study to perform 3D reconstruction, 300 4K images can take up to 10 hours to process, which is equivalent to analyzing only 5 seconds of video recording at 60 fps. In addition, each 4K image uses around 2.5-3 Megabytes of storage space. 10 minutes of video recording will, hence, use roughly 100 Gigabytes of storage space, which can accumulate over time. Furthermore, the number of images to be decomposed is also a function of the speed of movement. For instance, at instances with higher speed of movement, more fps is desired, compared to when the movement is still. Hence, utilizing all frames will be undesirable and impractical. In Section 3.1 a new definition of image overlap is provided, and a process is described to utilize the new definition of overlap to generate more frames at locations where the relative orientation between the two consecutive frames is larger and reduce frames at still locations.

*2.2. Photogrammetry in construction*

In the early utilization of images for progress monitoring in construction, a single pre-calibrated camera at a fixed location and orientation was adopted to report the level of completeness of building columns [32,33] as well as pre-cast concrete tunnel liners [34]. A single camera exposure is, however, insufficient to generate 3D point clouds (at least two exposures is necessary to recover depth [35]), which is desirable for dimensional quality checks. Furthermore, the presence of newly added elements on site can block the camera's field of view, resulting in an incomplete site coverage. Therefore, Golparvar-Fard [5,31,36], proposed to use multiple images in combination with SfM to generate dense 3D point clouds of construction sites for progress reporting. As explained in the previous section, SfM can only provide reconstruction up to an arbitrary scale factor. Therefore, to define the scale for metric reconstruction and to utilize the planned BIM for object recognition and progress reporting, the method proposed to use at least three carefully selected key-point correspondences between the point cloud and the planned BIM to perform registration through a similarity transformation (scale, rotation and translation [37]). At this stage, following a typical scan vs. BIM approach [38], it is customary to further refine the registration through iterative closest point matching between the closest BIM elements and the point cloud [39,40]. Once the point cloud and the BIM are registered and superimposed, the overlapping elements satisfying some similarity criteria (e.g. distance, percentage of coverage, colour, texture, and etcetera) are marked as complete [41]. In fact, particularly for photogrammetric point clouds, it is possible to also perform the point-to-model correspondence check at the image level through projecting the BIM onto the image plane using the camera EOPs and IOPs [5]. This latter property of images in combination with supervised learning methods has provided opportunities to enhance matching of image points with the BIM and consequentially improve progress reporting of different elements [42–45]. In addition to machine learning techniques, relationship-based reasoning [46] through work package sequences, and construction methods - which can be obtained from an appropriate BIM- can be utilized within a deep learning framework to further refine the results of scan vs. BIM, especially in the presence of occlusions [47,48].

So far it is established that the use of geometrical and elemental relationships, obtained from BIM combined with machine learning to label objects in images provides a strong tool for construction progress reporting. Especially when multiple elements with similar geometrical properties exist, a detailed BIM can provide additional means to facilitate with semantic feature extraction from point clouds [46]. The underlying assumption is, however, that an up-to-date BIM with sufficient level of development, is consistently available throughout the project's lifecycle. In addition, scan vs. BIM requires the model and the point cloud to be sufficiently close[5], and cannot incorporate the impact of construction errors [49] or changes in design specifications –that are not yet updated in the BIM– [4,46]. The use of supervised learning can of course increase the likelihood of matching even in the existence of larger construction errors; however, it still requires a large manually labeled image database of different construction materials and elements as training datasets, which may not be readily available [16,46]. Furthermore, inaccurate image to image registration (see Figure 2a below), and incorrect point cloud to BIM registration, in particular the scale definition, provide additional sources of error, which may impact the progress reporting accuracy. This manuscript, hence, assesses the feasibility for progress monitoring of pipes by:

1- Providing a process to adjust the number of images between specific video frames to optimize image to image matching and point cloud quality, while preserving computational efficiency.
2- A process for accurate scale definition using a low cost, lightweight, and portable black and white target board.
3- A method for automated detection of pipes from point clouds using only the geometric characteristics of cylinders, which is desirable when a 2D or 3D planned BIM does not exist.

## 3. Methodology

Following the previous discussions, the proposed methodology for automated progress reporting of mechanical pipes, using smartphone video recordings, consists of the following stages:

1- Optimizing the number of frames
2- Automated scale definition
3- Cylinder detection from point clouds
4- Classification of the detected cylinders based on their radii

The first two steps are designed to improve the overall photogrammetric point cloud quality through optimizing point density as well as defining accurate metric scale. Stages 3 and 4 are then proposed to enhance quality of pipe detection and classification of the point clouds. In the experiments, the impact of the first two steps for improving point cloud quality on the last two stages for enhancing pipe detection quality will be evaluated. These stages are explained in more detail in the following.

*3.1. Optimizing the number of frames*

In aerial photogrammetry, it is recommended to maintain a minimum overlie between consecutive images to achieve a reliable reconstruction [35]. With unmanned aerial vehicles (UAVs), it is possible to remotely control the speed of movement and plan the UAV's path in advance to achieve the desirable reconstruction accuracy and overlie between images (typically at least 80% overlie to account for wind). For indoor mapping with the existence of a planned BIM, it can be also possible to similarly plan an optimum mapping trajectory to increase point measurement accuracy as well as coverage [35,50], and to use an advanced mobile robot such as Spot [51] to control the position and orientation of movement at each instance. At present, until robots become mainstream and affordable on construction projects, a human worker must manually record the videos of the scene. Therefore, the position and orientation of the camera might be changed faster than ideal throughout the movement. At constant fps, it, hence, is possible to achieve a low image overlie between consecutive images, resulting in a gap. Figure 2a provides an example of the impact of drastic change of movement between frames on the registration of images. Figure 2a-left and 2a-right show the results of coarse 3D reconstruction using video frames decomposed into 0.5 fps and 2 fps, respectively. As illustrated by the blue oval in Figure 2a-left, a gap in the images appeared because of a relatively faster motion, causing a disparity in overlap as well as matched points between consecutive images, which in turn negatively impacted the registration between the images on the left and right of the gap. The impact of the poor registration on the dense 3D reconstruction can be observed in Figure 2a-bottom. It is known that reducing the base (distance between image views) can improve feature matching between different images [52]. This is since the number of features of a given image that is utilized for coarse reconstruction increases. Since the video recordings are of a continuous movement, the images in the vicinity of larger gaps are expected to contain lesser number of feature points contributing to the coarse reconstruction, compared to images with smaller gaps. In the following, a definition of image overlap in a network is provided that is used throughout the manuscript to help increase (remove) frames at instances of low (high) overlap.

---

[5] Especially, in cases where only BIM information and 3D point clouds are used.

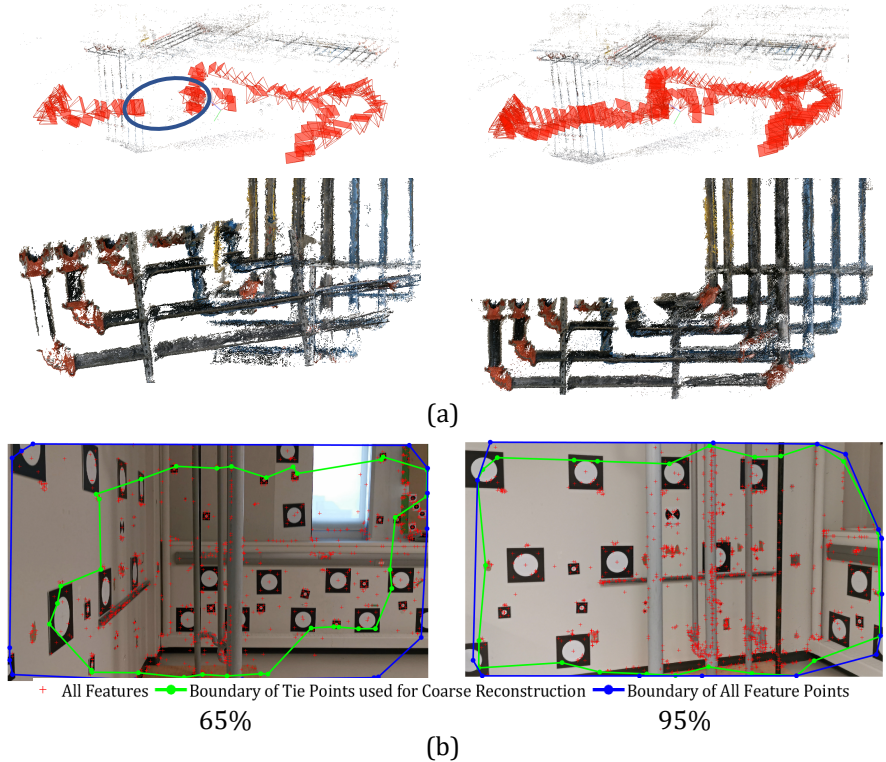

**Figure 2:** a) Impact of relatively faster position and orientation change between frames creating a gap at lower fps (left) and subsequently an error in reconstruction (bottom); b) examples of images with network overlap of 65% (left) and 95% (right)

*3.1.1. Definition of network overlap of images*

The degree of overlap of an image in the network, $Ov_N$, from here on, referred to as image overlap, is defined by:

$$Ov_N = \frac{A_{Ti}}{A_{Fe}} \qquad (1)$$

where $A_{Ti}$ is the area enclosed by all tie points of the image, used for coarse reconstruction, and $A_{Fe}$ is the area enclosed by all feature points of the image. Figure 2b shows an example of 65% and 95% overlap ratio for two images of the same network. Ideally, when all feature points are used within the coarse reconstruction process, the image is utilized to its fullest capacity. Conversely, if the image tie points only contain a fraction of all feature points, the image frame has not been utilized to its fullest potential and its effectiveness in the network can be improved. In the latter case, due to the continuous nature of video recordings (up to 60fps in our case), it is, hence, possible to increase the number of frames before and after an image to increase the number of matched feature points and subsequently increase the ratio of equation 1. In order to ensure each image overlap is larger than, $Ov_{min}$, and smaller than, $Ov_{max}$, **Algorithm 1: Optimizing Number of Frames**, is formulated as follows:

1- Start with an initial fps (0.5 fps is used in this study) and decompose the video into image frames.
2- Perform steps 1 through 7 of SfM, presented in Section 2.1, to estimate the EOPs of each image[6].
3- For each image, calculate the degree of overlap using equation 1.
4- Add frames satisfying the following:
    a. Find images whose degree of overlap is less than the minimum overlap, $Ov_{min}$.
    b. For each image satisfying step 4a, estimate the degree of similarity between its consecutive images (before and after). The degree of similarity between two images is directly correlated to the number of matched feature points between the images (see [53] for details).
    c. Add one frame between the two consecutive images of step 4b with the smallest similarity (lowest number of tie points).
5- Remove frames satisfying the following:
    a. Find images whose overlap is larger than the maximum overlap, $Ov_{max}$.
    b. From the images of step 5a, find groups of consecutive images with more than three images.
    c. Remove every even image from each group of consecutive images satisfying step 5b.
6- Perform the steps 2 through 5 until all images achieve the desirable overlap in the network.

In some cases, where considerable motion between consecutive frames exists, it is possible to encounter a phenomenon, called, motion blur. In such cases, it is recommended to deblur the image frame by first estimating the point-spread function (PSF), and then using a deblurring strategy, such as the Weiner deconvolution [53]. The PSF can be estimated through the evaluation of EOPs of consecutive frames from the SfM process. The aforementioned deblurring can be incorporated within Algorithm 1 for every newly added frame if there is a speculation of large motion between consecutive frames.

*3.2. Scale definition*

As explained in Section 2, SfM can provide 3D reconstruction up to an arbitrary scale factor, which is predominantly defined manually in construction literature [5]. Alternatively, black and white circular targets, which are commonly used in high precision metrology, can be utilized for registration and scale definition purposes [54]. The circular targets are approximated as ellipses in images and possess desirable properties such as a unique center, invariance under rotation and translation, and low cost of production [55]. Here, as shown in Figure 3, a black rectangular target board consisting of five white circles is designed. The real-world relative position and distance of the centers is known and measured a-priori. The targets can be utilized for scale definition purposes within the SfM framework if their 3D reconstructed centers are estimated automatically. Estimation of the centers of the target requires: (i) the EOPs and IOPs of the images in the field of view of the targets; (ii) the ellipses representing the targets in images; and (iii) the target centers in images. The EOPs and IOPs of the images are obtained using SfM (Figure 3a). The ellipses can be detected automatically via any reliable algorithm for ellipse detection from images. Here, the method presented in [56] is utilized (Figure 3b). The ellipse matching between images can then be carried out using Algorithm 2 of [26] (Figure 3c). Due to projective transformation, however, the actual centers of the circular targets do not coincide with the center of the best fit ellipses (see Figure 3d). This systematic error is referred to as the center eccentricity error, which must be corrected, especially in larger targets [55,57]. Current closed-form methods for correcting the eccentricity error [55,57] require detailed object space information such as target plane normal and actual target center to correct the eccentricity. In Maalek [26], a new method was proposed, which only requires the

---
[6] In case an image cannot be registered with the initial fps, increase one frame before and one frame after the image and perform step 2 again until all images are registered.

EOPs and IOPs of the images along with the ellipse geometric parameters in two views. The aforementioned discussions can be formulated using, **Algorithm 2: Scale Definition in SfM**, as follows:

1- Perform SfM to determine the EOPs and IOPs of the images (Figure 3a).
2- Detect ellipses using the ellipse detection of [56] (Figure 3b).
3- Match the ellipses representing targets between different views using Algorithm 2 of [26] (Figure 3c).
4- Adjust the ellipse's center for systematic eccentricity using Algorithm 3 of [26] (Figure 3d).
5- Perform triangulation to determine the 3D coordinates of the centers.
6- Use the following equation to define the scale [37]:

$$s = \left(\frac{\sum_{i=1}^{k}(P_i-\bar{P})^2}{\sum_{i=1}^{k}(Q_i-\bar{Q})^2}\right)^{\frac{1}{2}} \tag{2}$$

where s is the scale, $k$ is the number of circular targets detected, $P_i$ and $Q_i$ are the object space coordinates, and 3D reconstructed coordinates of the targets, respectively, and $\bar{P}$ and $\bar{Q}$ are the average of $P_i$ and $Q_i$, respectively.

The impact of the proposed algorithm for scale definition, in particular the target center adjustment, is still unknown. In addition, for practical applications, it is desirable to determine the minimum number of images to be captured from the target board to achieve an accurate 3D reconstruction. The effects of center adjustment along with the required number of images are investigated in the experiments.

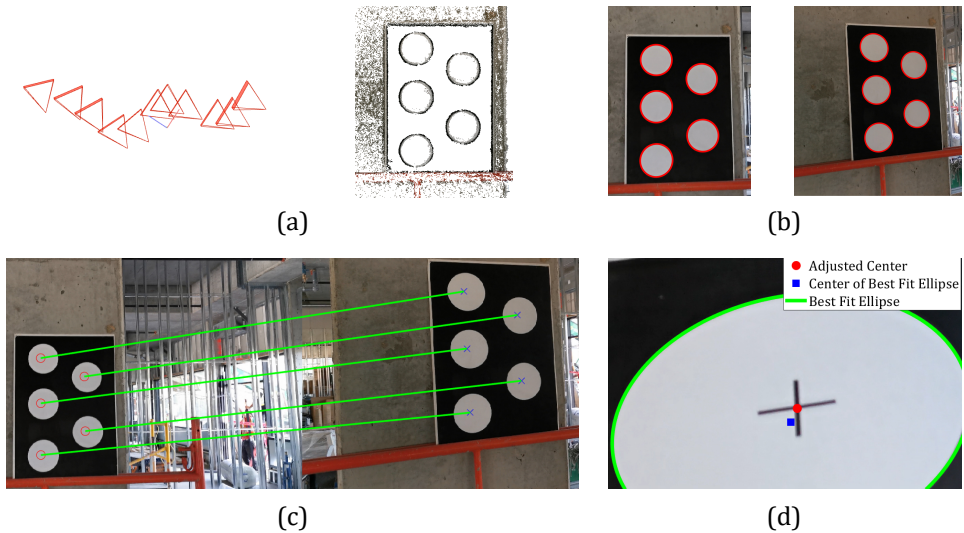

**Figure 3:** Process used to define scale using the black and white target: a) EOPs of target images (left) and 3D reconstructed target point cloud; b) detected ellipses (shown in red) from images; c) matching ellipses using the estimated EOPs; and d) adjusted center of the best fit ellipse

*3.3. Cylinder detection from point clouds*

In Maalek [4], an efficient and robust method was presented to extract a single cylinder from an isolated set of point clouds. The points of interest for each cylinder were, however, isolated using the planned BIM. Here, to provide an initial set of points to be considered for cylinder detection, a unique property from the intersection of planes and cylinders is utilized. It is well-known that the intersection of a cylinder and a plane forms an ellipse [58]. More specifically, if a random plane is intersected with the point cloud of a cylinder, the parallel projection of the points onto the plane in the vicinity of the plane are expected to form an ellipse. The cylinder's parameters can then be recovered using the following formulae [58]:

$$\begin{cases} R = b_e \\ \delta = \cos^{-1} \frac{b_e}{a_e} \\ C_c = Rot^T . C_e \end{cases} \quad (3)$$

where $R$ and $C_c$ are the radius, and the point of intersection between the cylinder's axis and the plane, respectively, $C_e = (x_e, y_e, z_0)^T$, $a_e$ and $b_e$ are the center, semi-major length and semi-minor length of the ellipse in the coordinate system of the intersecting plane, respectively, $Rot$ is the rotation matrix that transforms the plane's normal vector to vector $(0,0,1)^T$, and $\delta$ is the angle between the plane's normal vector and the cylinder's axis. Therefore, given a random plane passing through cylindrical point clouds, the problem of cylinder detection in 3D can be reduced to ellipse detection in 2D. Once the ellipses are detected, an initial estimate of the cylinder's axis, radius and center can provide a set of isolated points of interest to be fed to the robust cylinder detection of [4]. To acquire the cylinder's axis from equation 3, however, the normal vector must be rotated either $+\delta$ or $-\delta$ along the semi-minor axis [59] (two-fold ambiguity). Therefore, both possible solutions are first used to generate isolated interest points for each cylinder. Given a random plane passing through the cylindrical point clouds (Figure 4a), **Algorithm 3: Cylinder Detection from Projected Point Clouds**, is developed as follows:

1- Find points on the point cloud within $d = 1cm$ of the given plane.
2- Rotate the plane such that the normal to the plane is parallel to the $z$ axis.
3- Perform connected components region growing in the $x - y$ plane using Algorithm 5 of [4] (Figure 4b).
4- Detect ellipses using the non-overlapping ellipse detection of [56,60] (Figure 4c).
5- For each detected ellipse perform the following:
    a) Estimate $\delta$, $R$, $C_c$ using equation 3.
    b) Rotate the random plane's normal vector $+\delta$ or $-\delta$ about the minor axis of the ellipse.
    c) For each possible cylinder axes of step 5b, perform the following:
        1. Find the rotation matrix, $Rot_{az}$, that transforms the axis of step 5b parallel to the $z$-axis.
        2. Determine all points within $\gamma R$ from the rotated center of the cylinder, $Rot_{az}.C_c$, in the $x$-$y$ plane. Following [4], the cylindrical radius is defined as: $\gamma R = \max(2R, R + 10cm)$.
        3. Find the root mean squared error (RMSE) of the inlier points following a circular pattern using Algorithms 1 and 2 of [4].
    d) The cylinder's axis achieving the lowest RMSE (step 5c3), among the two candidates, is considered as the cylinder's initial axis.
    e) Perform the robust cylinder axis estimation, Algorithm 3 of [4], on the points obtained by step 5c2 using the initial cylinder axis to estimate the parameters of the best fit cylinder (Figure 4d).

Given that mechanical pipes are commonly attached from one side to a wall, the reconstructed point cloud of the pipes are only partially visible. The proposed method in Algorithm 3 can handle occlusions, partial visibility, and noise due to the following:

1- Step 4 and 5a uses the method of [60] to fit ellipses to partially occluded elliptic edge points. The method was shown to estimate the radius, center, and axis of the partially visible parent cylinder with varying noise levels within 3.4mm, 4.7mm, and 3.9°, respectively, for 30,000 different configurations.
2- Step 5 then utilizes the robust cylinder detection of [4], which was shown to produce very reliable cylinder detection using initial cylinder axis approximations within 7° in occluded, partially visible, and noisy point clouds in the presence of outliers [4].

The combination of these two steps is, hence, expected to produce reliable results in partially visible cylinders as will be verified and validated in the presented experiments.

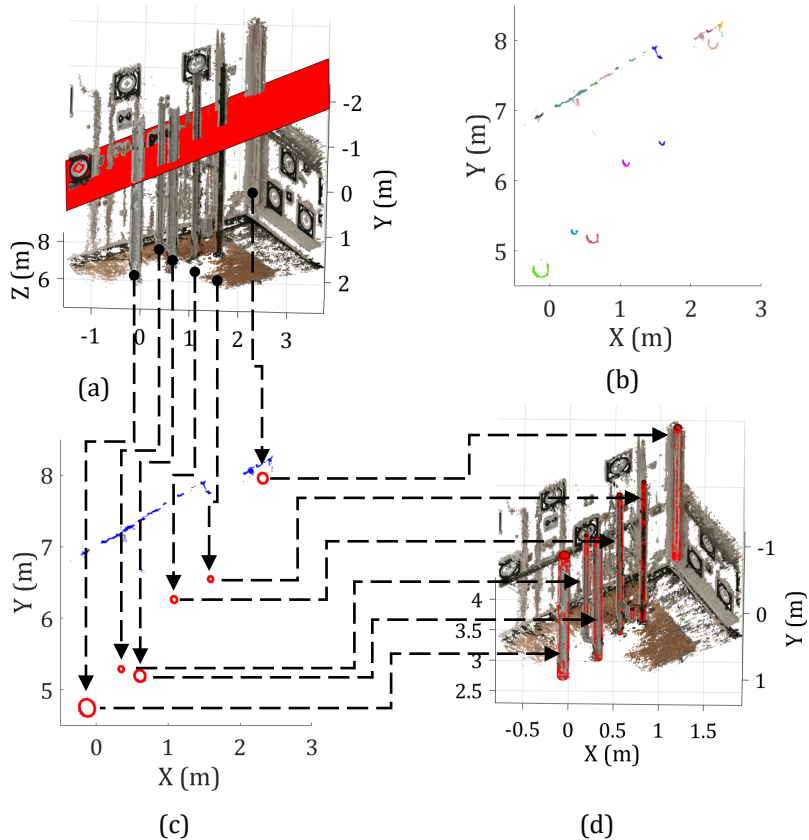

**Figure 4:** Procedure for cylinder detection from point clouds: a) plane intersected with point cloud; b) connected components region growing on 2D parallel projected points onto the image; c) detected ellipses (shown in red) from each connected segment; and d) detected cylinders superimposed onto the point cloud.

The last consideration for Algorithm 3 is the generation of the initial plane. At this point, this plane is defined semi-automatically. In practice, this can be accomplished semi-autonomously by: (i) drawing a line (say by swiping on the smartphone's touchscreen) that passes through the pipes on at least one video frame; (ii) determining the object-space 3D coordinates of at least three random points on the considered line (the 3D points must be non-collinear); and (iii) estimating the best fit plane passing through the 3D coordinates of the points.

### 3.4. Radius-based classification of cylinders

In this section, the cylinders are classified based on their radii, which is estimated using Algorithm 3. The information regarding the outer radius/diameter of planned/ordered pipes installed are readily available from the pipe schedules found on the bill of quantities. Therefore, the problem is to cluster the detected cylinders into groups of pipes with known outer radii. This is a typical k-means clustering problem [61,62]. Here, the planned outer radii of pipes are used as the initial means of each cluster. The result of the k-means clustering is schematically shown in Figure 5.

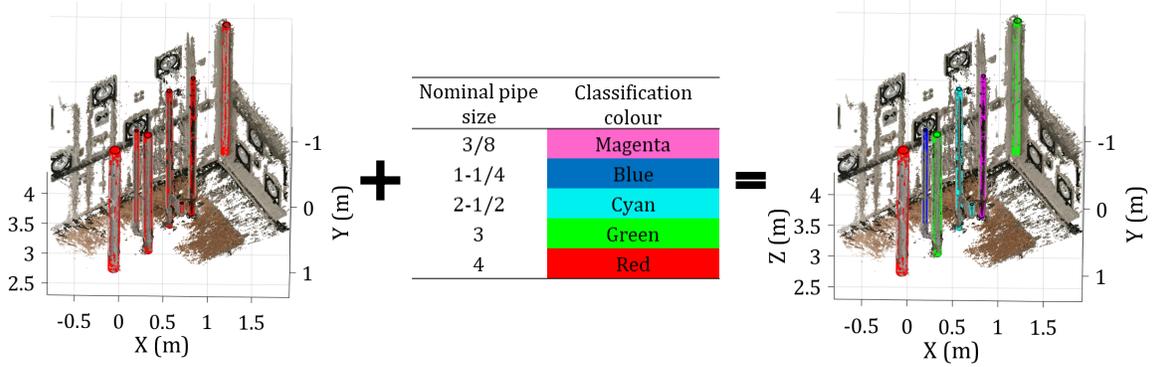
**Figure 5:** k-means clustering for classification of detected cylinders based on radius

*3.5. Metrics for validation of results*

Three main categories of metrics were used in this study to assess the effectiveness of the proposed methods, namely accuracy of radius estimation, percent error of length change, and the quality of cylinder classification with respect to the ground truth information. The accuracy of radius estimation is defined here as the root mean squared error (RMSE) from the ground truth as follows:

$$RMSE = \sqrt{\frac{1}{N}\sum_{i=1}^{N}(r_i - r_{True})} \qquad (4)$$

where $r_i$ and $r_{T_i}$ are the estimated radius and ground truth radius of cylinder $i$, and $N$ is the number of cylinders considered for the estimation of accuracy. The ground truth radius of a pipe is determined from the outer diameters specified from the pipe's manufacturer. The percentage error (or percentage of relative error), $\delta$, of the estimated length of a given pipe is defined using the following equation:

$$\delta = \left|1 - \frac{L_{Estimated}}{L_{True}}\right| \times 100 \qquad (5)$$

where $L_{Estimated}$ is the estimated length of a detected pipe as per the convention specified in [46] for cylinders, and $L_{True}$ is the ground truth length, determined manually. The last metrics, the precision, recall, accuracy, and F-measure [63], are utilized to measure the quality of cylinder classification:

$$\begin{cases} Precision &= \frac{T_P}{T_P+F_P} \\ Recall &= \frac{T_P}{T_P+F_N} \\ Accuracy &= \frac{T_P+T_N}{T_P+T_N+F_P+F_N} \\ F-measure &= 2.\frac{Precision.Recall}{Precision+Recall} \end{cases} \qquad (6)$$

where $T_P, T_N, F_P, F_N$ are the number of true positive, true negative, false positive, and false negative of the detection, respectively. The $T_P, T_N, F_P, F_N$ were determined based on the correct number of cylinders of a particular type and radius from the planned construction information.

## 4. Experiment design

Four sets of experiments were designed to assess the effectiveness of the proposed methods. The first three experiments were carried out in controlled laboratory conditions, and the last experiment was carried out in an actual construction site. The first experiment is related to the impact of number of image views on the proposed method for scale definition (Algorithm 2). The second experiment evaluates the impact of the

Page 12 of 26

proposed image overlap metric on point cloud quality and in particular accuracy. The third set of experiments assess the impact of changing the image network overlap (equation 1 and Algorithm 1) on the accuracy of radius, percent error of length, and the quality of classification of the detected pipes. The last experiment involves the steps taken in the third experiment, repeated on an actual construction site. For all experiments, 4K videos were captured using a pre-calibrated Huawei P30. In this manuscript, the smartphone cameras were pre-calibrated at the calibration laboratory (see Figure 6a), which includes a large set of black and white circular target field, using the methods described in [26,27].

## 4.1. Laboratory and construction site details

The laboratory included six mechanical mock pipes that were professionally installed in the corner of a complete calibration room (Figure 6a). These pipes are utilized for the controlled laboratory experiments presented in this study. The mechanical mock pipes were of five different outer radii, consisting of polyvinyl chloride (PVC) sanitary pipes and steel heating/cooling pipes to simulate the variety that exists on a typical construction site. The field experiments were carried out at an indoor construction site with 58 pipes in visibility of the recorded videos. The 58 pipes consist of six classes of outer diameters, with only one larger class of pipe added to the five classes of the laboratory experiments. A picture of the portion of the construction site under study is presented in Figure 6b. For the reference of the reader, the average linear speed of movement of the recorded video footage of the laboratory and the construction site were 0.78m/s and 0.72m/s, respectively. The angular speed of movement for the laboratory and construction site were 3.6°/s and 4.5°/s, respectively.

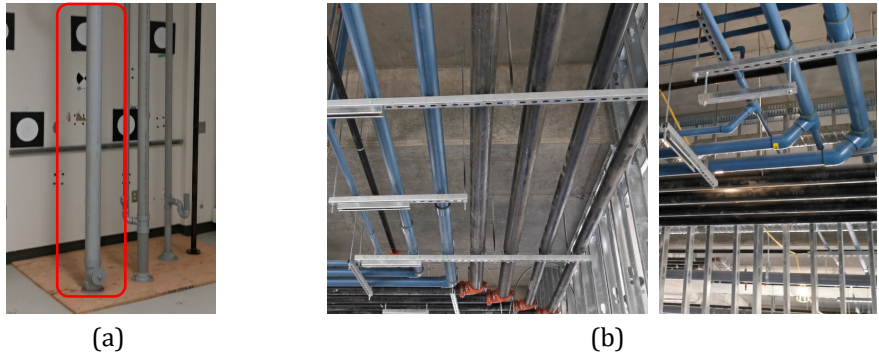

(a)          (b)

**Figure 6:** Images of the experiment setup: a) pipes in calibration laboratory; and b) mechanical pipes in indoor construction site

## 4.2. Laboratory experiment 1: scale definition vs. image views

This experiment was designed to quantify the impact of the proposed center adjustment on the scale definition as the number of image views increase. It is hypothesized that as the number of image views increases, the error associated with scale definition reduces. Quantifying the relationship between the image views and the scale definition accuracy enables the determination of the minimum number of views required to achieve a desirable scale definition accuracy. To this end, the proposed target board (Figure 3) is attached to the wall in the background of one of the professionally installed mock pipework (shown with red rectangle in Figure 6a). 35 seconds of video recordings were captured from the pipe and background target board. The video was decomposed at 1fps. The EOPs of the 35 images using SfM is shown in Figure 7a-left. All 35 images were used to generate the dense 3D reconstruction of the pipe. The radius of the pipe in the arbitrarily defined scale is estimated using Algorithm 3. The impact of the proposed scale definition (Algorithm 2) with the adjusted centers and without center adjustment (best fit ellipse center) on the accuracy of the estimated radius of the pipe was examined as the number of image views increased from 2 to 35. Since more than one combination of image views exists, for a given number of (say $k$) image views, $N$ different combinations of $k = 2 \ldots 35$ images were selected. To this end, for $k$ image views, the following steps are carried out:

1- Randomly select $N$ different combination of $k$ images from the 35 images.
2- For each set of $k$ images, define the 3D reconstructed scale with Algorithm 2 once using the estimated best fit ellipse center of the targets (no adjustment), and again using adjusted target centers.
3- Estimate the radius of the pipe using both defined scales in step 2.
4- Calculate the accuracy of the pipe's radius using equation 4 for both scales.
5- Record the mean of the radius accuracies, obtained from step 4 for the $N$ combinations of $k$ images.

In this experiment, $N = 50$ combinations are used.

*4.3. Laboratory experiment 2: point cloud quality vs. image overlap*

There are two main accepted predictors of point cloud quality, namely, density and accuracy [29,64–66]. Increasing the image overlap will increase point cloud density as shown in Tables 1 and 2 as well as Figure 7. Considering the known impact of image overlap on point cloud density, the focus of this experiment will be to evaluate the impact of the defined image overlap on point cloud accuracy. To this end, the convention proposed in [66] to evaluate the accuracy of building point clouds, acquired from smartphone cameras, is adopted. In this approach, the accuracy is defined by point-to-point [65] correspondence matching between registered photogrammetric (source) and TLS (reference) point clouds. The following steps were carried out for image overlaps increasing from 70% to 95% in 5% increments to implement the strategy:

1- Register the photogrammetric (source) and TLS (reference) point clouds using a similarity transform [37]. Since the experiments were carried out at the calibration laboratory, the automatically acquired centers of 10 targets were used as point correspondences between the point clouds.
2- Find the nearest neighbor of the registered source point cloud from the reference point cloud to find the point correspondences. Here, only points, whose distance from its closest point were within 10cm, are considered.
3- Record the mean of the distances from step 2 for each image overlap.

*4.4. Laboratory experiment 3: pipe classification vs. image overlap*

This experiment was designed to assess the effectiveness of the proposed method for classification of pipes based on their radii as a function of image overlap. To this end, the minimum image overlap was increased from 70% to 95% in 5% increments. The maximum image overlap for each configuration was set as 2.5% more than the minimum overlap. The summary of the reconstruction results for each image overlap percentage is provided in Table 1. As illustrated, as the image overlap percentage is increased, the number of tie points as well as dense reconstruction, increases. This can also be visually observed from the reconstruction results for 70%, 80%, 90%, and 95% overlap, presented schematically in Figure 7b. For each image overlap, Algorithm 1 was used to determine the images, which were used for dense 3D reconstruction. The scale was defined using Algorithm 2 and the pipes were detected using Algorithm 3. Each detected cylinder was then classified based on the true pipe radius as described in Section 3.4. For each image overlap, the accuracy of the estimated radius, percent error of the estimated length, and the classification quality (using equations 4 through 6) for the detected pipes were reported.

**Table 1:** Laboratory experiment setup: results of the dense 3D reconstruction for each image overlap

| Number of images used | Image overlap (%) | Number of tie points | Dense reconstruction (# of points) |
|---|---|---|---|
| 10 | 70% | 1,084 | 79,963 |
| 14 | 75% | 2,299 | 220,942 |
| 18 | 80% | 2,900 | 291,176 |
| 25 | 85% | 4,535 | 438,667 |
| 48 | 90% | 6,963 | 723,973 |
| 70 | 95% | 12,578 | 976,549 |

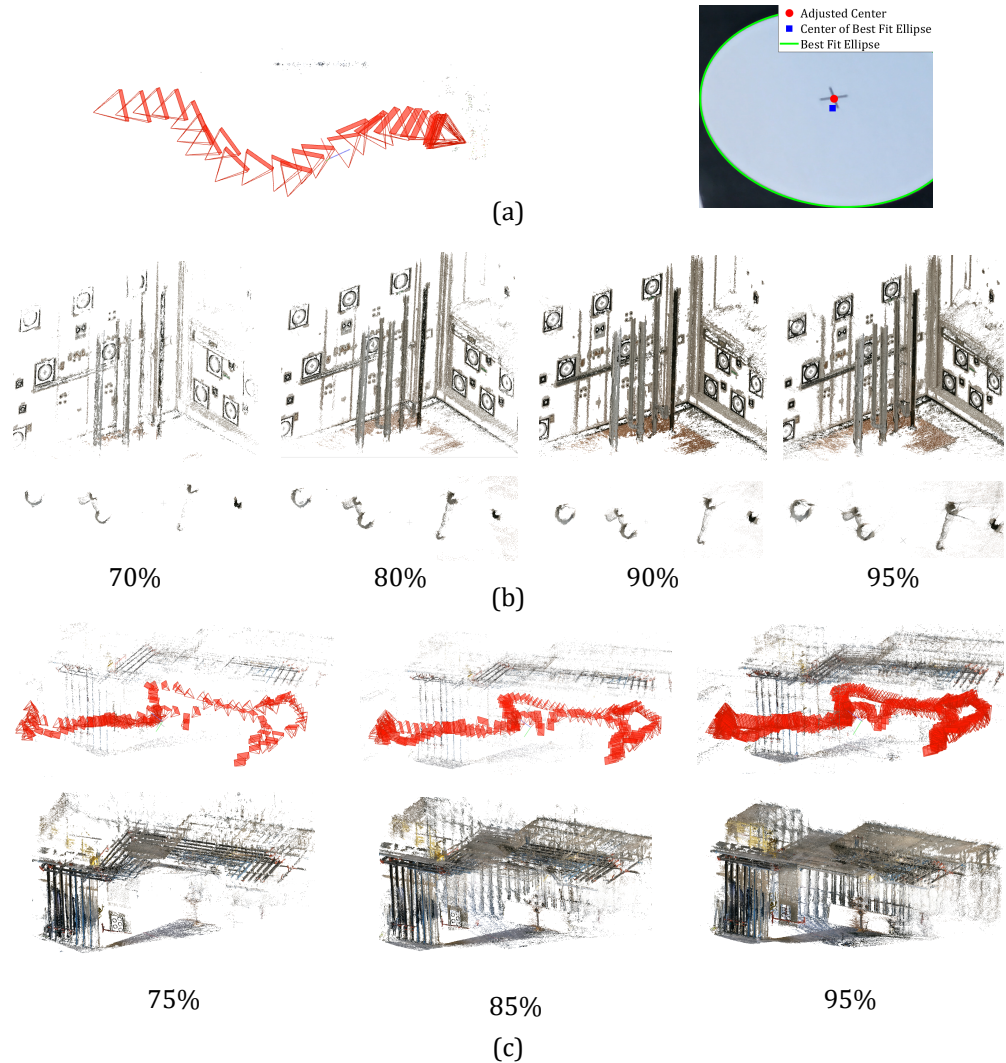

(a)

70%    80%    90%    95%
(b)

75%    85%    95%
(c)

**Figure 7:** Schematic presentation of the experiment setups for: a) scale definition as a function of image views; b) laboratory pipe detection and classification experiment; and c) field experiment

*4.5. Field experiment: pipe classification vs. image overlap*

This experiment was designed to determine the effectiveness of the proposed methods in a larger environment under real construction site conditions. The previous experiment, presented in Section 4.4, was, hence, repeated for the data captured from an indoor site during construction with the exception that the desired overlap was assessed at three instances with minimum overlap of 75%, 85% and 95%. Furthermore, the portion of the site under study consists of 58 pipes, categorized into six classes based on their outer radii. Since many pipes of a specific radius exist, the accuracy, and relative error were also reported as a function of the classes of pipe radius. To maintain consistency between the laboratory and the construction site, the video was recorded at an average distance of approximately 2m from the surface of the pipes. The summary of the dense reconstruction, including number of tie points used for coarse reconstruction, are reported in Table 2. The generated images using Algorithm 1 for the three settings as well as the final dense reconstruction is shown in Figure 7c. As illustrated, similar to the laboratory experiment, the number of tie points as well as the final dense reconstruction have increased as the image overlap requirements were increased.

**Table 2:** Field experiment setup: results of the dense 3D reconstruction for each image overlap

| Number of images used | Image overlap (%) | Number of tie points | Dense reconstruction (# of points) |
|---|---|---|---|
| 133 | 75% | 38,278 | 2,527,411 |
| 177 | 85% | 103,214 | 4,630,226 |
| 355 | 95% | 250,589 | 11,586,293 |

## 5. Experiment results

### 5.1. Laboratory experiment 1: scale definition vs. image views

The number of image views were increased from 2 to 35, and the accuracy of the radius estimation of one professionally installed mock pipe was measured using scale definition (Algorithm 2) with center adjustment, and without center adjustment. Figure 8 shows the results of the mean of the 50 simulations as a function of the number of image views. Three important observations can be made from Figure 8. First is that the estimated radius using the scale definition with the adjusted target center, consistently outperformed that obtained using the best fit ellipse centers (improvement by a factor of around 2 on average). The second observation was that observing the target field with only two views is considerably less accurate than when the target is observed by three views. This demonstrates that targets should at least be observed by three views to achieve sub-millimeter pipe radius estimation accuracy. The last observation was that for both adjusted and unadjusted centers almost remained constant when the number of image views increased from 5 to 35 images (around only 0.02mm standard deviation). Therefore, 5 image views can be considered sufficient and can be recommended to produce accurate scale definition for dense 3D reconstruction.

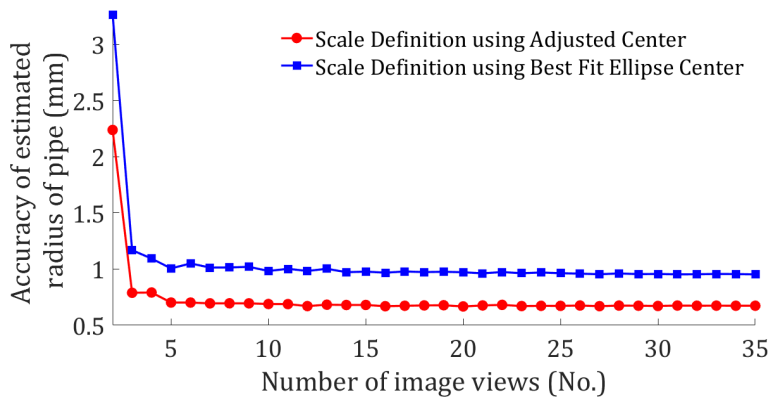

**Figure 8:** Impact of scale definition (Algorithm 2) with target center adjustment, and without center adjustment (best fit ellipse) as a function of the number of image views

*5.2. Laboratory experiment 2: point cloud quality vs. image overlap*

The accuracy of photogrammetric point clouds compared to the reference TLS point cloud was evaluated as the image overlap increased from 70% to 95% in 5% increments. Figure 9 shows the heat map of the distance between the photogrammetric and TLS point clouds for four samples with image overlaps of 70%, 80%, 90%, and 95% as a point of reference. Figure 10 shows the RMSE of the distances between the photogrammetric and TLS as the image overlap increases. Since photogrammetric point clouds will produce larger number of points as the image overlap increases (see Tables 1 and 2), the results of two configurations were reported in Figure 10. The first (shown in blue with square markers) represents the RMSE using all data points. The second, shown in red with rounded markers, presents the RMSE for the closest points at each image overlap to the 79,963 points (see Table 1), obtained from dense reconstruction using the 70% image overlap. The latter minimizes the impact of the number of points on the results and compares the quality of the same points between different image overlaps to provide a fair basis for comparison of RMSE. It was observed that the RMSE appeared to improve as the image overlap increased for both configurations. When considering all points, the difference in the RMSE between 70% and 95% was less than 1mm. However, when the same points between different image overlaps were considered, the difference in the RMSE between the 70% and 95% overlaps increased more considerably to around 2.3mm. The results of the experiment suggest that increasing the image overlap improves both point cloud density, and accuracy; hence, image overlap can be considered as a good predictor of point cloud quality.

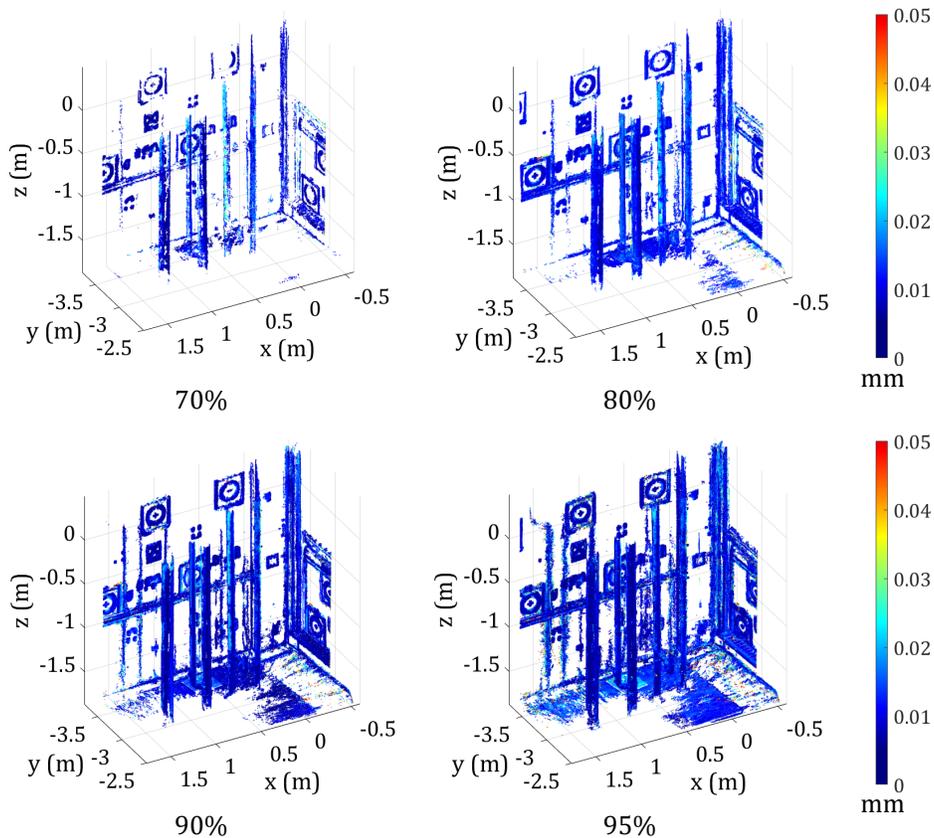

**Figure 9:** Heat map of the local distance (in mm) between the photogrammetric and TLS point clouds for image overlaps of 70% (left); 80% (middle-left); 90% (middle-right); and 95% (right)

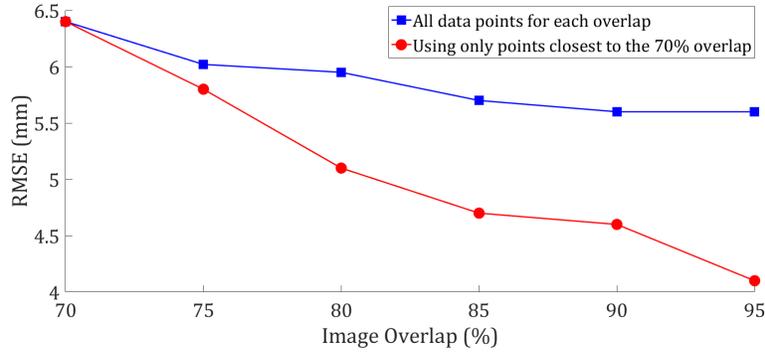

**Figure 10:** RMSE of the photogrammetric and TLS point clouds as a function of image overlap using all point clouds (blue-square marker), and only points closest to the those of the 70% overlap (red-round marker)

*5.3. Laboratory experiment 3: pipe classification vs. image overlap*

The quality of cylinder extraction, accuracy of radius estimation, and percentage of relative error of measured length are determined as the image overlap increases from 70% to 95% in 5% intervals. Table 3 shows the detection quality for each image overlap. It can be observed that as the desired percentage of image overlap increases in Algorithm 1, the overall quality of cylinder extraction (F-measure) also improves. This is mostly due to the increased point density, which increases the likelihood of the ellipse detection from a set of parallel projected 2D points onto the intersecting plane. The possible impact of point density on the cylinder detection was also mentioned in [67]. The 90% and 95% overlap provided perfect F-measure for the cylinder detection and classification, demonstrating that sufficient point density was attained. It is to mention that experimentation with more pipes is required to draw any conclusive remarks, particularly regarding the accuracy of the pipe classification, which will be treated in the field experiments.

**Table 3:** Laboratory experiment- results of the pipe extraction quality as a function of image overlap

| Image Overlap (%) | Precision | Recall | Accuracy | F-measure |
|---|---|---|---|---|
| 70% | 57.1% | 80.0% | 75.0% | 66.7% |
| 75% | 71.4% | 83.3% | 85.0% | 76.9% |
| 80% | 71.4% | 83.3% | 87.0% | 76.9% |
| 85% | 85.7% | 85.7% | 91.3% | 85.7% |
| 90% | 100.0% | 100.0% | 100.0% | 100.0% |
| 95% | 100.0% | 100.0% | 100.0% | 100.0% |

Table 4 shows the accuracy of the estimated radius along with the relative error of the estimated length. To provide a fair comparison, only the results of the correctly matched cylinders are reported for each image overlap. As illustrated, even for the correctly matched cylinders, 70% overlap appears insufficient to provide sub millimeter accuracy. Furthermore, the relative error of the pipe length barely supersedes 10% for image overlaps of 80 % and 85%. The results for the case with 90 % and 95% appear to be more stable and almost identical with the pipe length achieving a slight improvement in the 95% case. The last observation was that both the radius accuracy as well as the length change improved as the desired image overlap increased. This is most likely a result of: (i) higher point density in larger image overlaps, generating more redundant points, which can help improve the cylinder fitting accuracy; and (ii) larger number of tie points that were used in larger image overlaps, providing additional redundancy to estimate the EOPs of images with greater accuracy.

**Table 4:** Laboratory experiment- results of the accuracy of the estimated pipe radius along with the relative error of the pipe lengths as a function of image overlap

| Image overlap (%) | Accuracy of estimated radius (mm) | Relative error of the estimated length (%) |
|---|---|---|
| 70% | 13.4 | 15.3% |
| 75% | 9.0 | 12.9% |
| 80% | 7.9 | 9.8% |
| 85% | 6.5 | 9.1% |
| 90% | 5.1 | 5.9% |
| 95% | 5.1 | 5.5% |

*5.4. Field experiment: pipe classification vs. image overlap*

The portion of the indoor field under study, consisted of 58 pipes in six classes of pipe diameters. The extracted pipes for the six classes are shown in Figure 11. Here, to observe the impact of the image overlap on different pipe classes, the accuracy of the estimated radius along with the relative length deviation is reported for each pipe class. Table 5 shows the quality of cylinder detection and classification. Similar to the laboratory experiments, both Type I (precision) and Type II (recall) improved as the image overlap was increased from 75% to 95%. In the laboratory experiment, however, the precision of 100% was achieved, whereas the field experiments achieved a precision of 93.1% at 95% overlap. Upon further examination, it was found that 4 pipes of the smallest diameter at relatively farther distances (about 3m compared to the total average of 2m) were not identified using our method. This was since the point density of these pipes were considerably less than the closer pipes with larger diameters. This demonstrates that even though the percentage of overlap provides a basis to increase the number of images to improve its effectiveness, another metric is required to further increase the number of required images in the vicinity of frames observing relatively farther points.

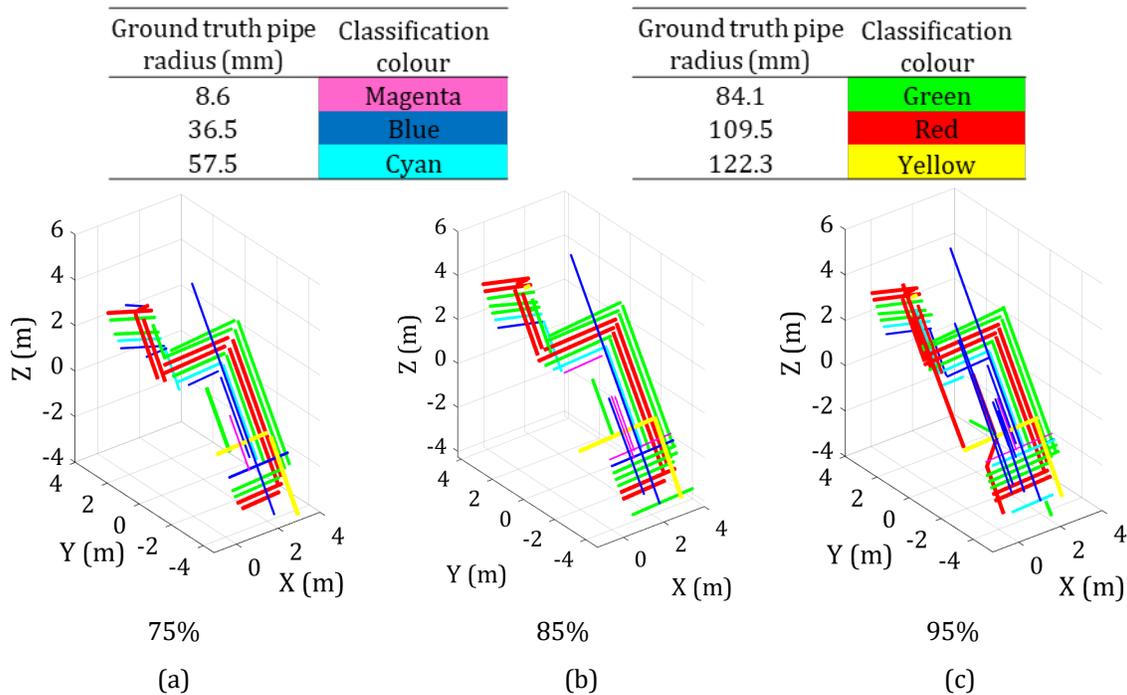

| Ground truth pipe radius (mm) | Classification colour |
|---|---|
| 8.6 | Magenta |
| 36.5 | Blue |
| 57.5 | Cyan |
| 84.1 | Green |
| 109.5 | Red |
| 122.3 | Yellow |

(a) 75%  (b) 85%  (c) 95%

**Figure 11:** Results of the classification of the 58 pipes into the colour-coded six classes as a function of the desired image overlap of: a) 75%; b) 85%; and c) 95%

**Table 5:** Field experiment- results of the pipe extraction quality as a function of image overlap

| Average Overlap (%) | Precision | Recall | Accuracy | F-measure |
|---|---|---|---|---|
| 75% | 60.0% | 89.2% | 71.1% | 71.7% |
| 85% | 74.5% | 91.1% | 80.0% | 82.0% |
| 95% | 93.1% | 100.0% | 95.7% | 96.4% |

Tables 6 and 7 demonstrate the impact of the image overlap on the pipe radius estimation as well as the length estimation for the different pipe classes. As illustrated from the overall results, both the radius estimation accuracy and the relative error in the length estimation improved as the image overlap increased. From Table 6, it was observed that the radius estimation accuracies for the smallest pipe was impacted more by the increase in the image overlap, than say the largest pipe. This is most likely attributed to the lower point density of the pipes at lower overlap percentages, which will generally impact smaller objects and pipes more severely. From the results presented in Table 7, the relative error of pipe's length shows improvement as the image overlap increases; however, a particular pattern or trend as a function of the size of the pipes was not observed.

The final observation is that the laboratory and field experiments produced relatively consistent results in terms of the impact of image overlap on classification quality, radius estimation accuracy, and length percent error accuracy. By comparing Tables 3 and 4 with Tables 5 through 7, the average percent error of the F-measure, radius accuracy, and length change of the correctly classified pipes, between the laboratory and field were approximately 5%, 7% and 9%, respectively. In addition, the classification quality, radius estimation and length change appeared to improve as the image overlap increased for both laboratory and field experiments.

**Table 6:** Field experiment- results of the accuracy of the estimated pipe radius as a function of image overlap

| Ground truth outer radius of pipe (mm) | Accuracy of radius estimation (mm) | | |
|---|---|---|---|
| | 75% | 85% | 95% |
| 8.6 | 18.2 | 14.5 | 7.8 |
| 36.5 | 8.3 | 6.9 | 6.6 |
| 57.5 | 8.9 | 7.7 | 3.9 |
| 84.1 | 9.1 | 8.1 | 6.1 |
| 109.5 | 8.9 | 5.8 | 4.0 |
| 122.3 | 7.1 | 6.9 | 6.4 |
| Overall | 9.1 | 7.5 | 5.4 |

**Table 7:** Field experiment- results of the percent error of the estimated pipe length as a function of image overlap

| Ground truth outer radius of pipe (mm) | Relative error of the length estimation (%) | | |
|---|---|---|---|
| | 75% | 85% | 95% |
| 8.6 | 10.2% | 10.8% | 8.0% |
| 36.5 | 8.6% | 7.2% | 3.3% |
| 57.5 | 9.3% | 8.9% | 7.8% |
| 84.1 | 12.8% | 10.4% | 6.5% |
| 109.5 | 12.8% | 8.4% | 3.8% |
| 122.3 | 8.0% | 2.3% | 2.4% |
| Overall | 11.2% | 8.6% | 5.0% |

## 6. Conclusions

This manuscript presented a new approach towards the automatic progress reporting of mechanical pipes on construction projects using smartphone cameras, based on estimating the lengths of installed pipes, classified with respect to their planned radii. The method extracts cylinders from point clouds, obtained from dense reconstruction of video recordings. The cylinders are then classified into existing classes of pipes (acquired from the pipe schedules in the bill of quantities), based on their outer radii. Since the video recordings are captured by a human worker with possible inconsistent movements (e.g. speed), a process was developed and a new definition for image overlap was proposed to remove redundant frames and to increase the frame rates where required. The larger image overlaps were expected to provide more densely reconstructed points, which in turn was speculated to help increase the possibility to correctly detect more cylinders from the point cloud (this was later corroborated from both the laboratory and field experiments). The manuscript also provided an automated procedure for accurate scale definition in 3D reconstruction, which is specifically important for quality assessment and in our case for classification of pipes based on their real-world radii.

To assess the effectiveness of the proposed methods, four sets of experiments at both a controlled laboratory setting as well as a real-world construction site were designed. The first experiment was designed to assess the requirements for scale definition, in particular the number of image views required to cover the proposed target field, and the necessity to adjust the systematic eccentricity error of the centers of the targets. It was shown that the scale defined by adjustment of the eccentricity errors –created as a consequence of the projective geometry of circular targets in images– improved radius estimation accuracy (by a factor of 2 on average), compared to when the centers of the targets were not adjusted for eccentricity. It was also observed that five views were sufficient to provide an accurate scale definition to estimate the radius of a pipe with sub-millimeter accuracy.

The second experiment was developed to evaluate the impact of increasing the image overlap on the point cloud quality, and more specifically accuracy. It was shown that increasing image overlap not only improves point cloud density, but it also enhances the point cloud accuracy. Image overlap can, hence, be considered as a good predictor of photogrammetric point cloud quality, particularly in cases where other means of assessing the point cloud quality through external measures is not available (e.g. when no reference TLS data is available for comparison).

The third experiment was designed to determine the accuracy of the radius estimation, the percentage error of the length estimation and the classification quality of the detected cylinders as a function of image overlap in controlled laboratory settings. Six mechanical mock pipework were professionally installed in a controlled laboratory setting, and the minimum image overlap was changed from 70% to 95% in 5% increments. It was observed that the quality of detection, radius estimation accuracy as well as percent error of length improved consistently as the image overlap increased. It was also observed that the results obtained from 90% and 95% overlap were almost synonymous.

The last experiment involved the repetition of the laboratory experiment on an actual construction site with 58 installed mechanical pipes in the field of view of the smartphone video recording. Overall, it was observed that the field experiments achieved similar results in terms of average radius estimation, length change and classification quality of detected pipes, compared to the laboratory experiments. The results imply that the increased scale of the experiments between the laboratory and field (from six pipes to 58 pipes) did not significantly impact the quality of classification, accuracy of radius estimation, and percent error of the estimated length. It was also observed that lower overlaps perform worse for pipes of smaller diameters not only in terms of classification quality, but also in terms of radius and length change estimation of correctly classified pipes. This shows that the increased image overlap, not only improves the cylinder extraction quality through increase in point density (which was expected [59]), but it also enhances the accuracy of the estimated parameters.

The methods presented in this manuscript showed promise toward the automatic extraction of length of the pipe as a means of reporting the progress of mechanical works on construction projects. The following provide additional avenues to expand and improve upon the proposed work in the future:

1- The proposed cylinder detection method requires the definition of an initial plane that must pass through the cylinders. Currently, this plane is anticipated to be defined semi-automatically. Development of new methods to define this plane automatically or to use supervised leaning and deep learning to extract the location of the pipes in each image will provide interesting avenues for further exploration.
2- The proposed pipe classification method classifies pipes based on their radii. However, pipes of different types (e.g. sanitary vs. heating), and material (e.g. plastic vs. steel), may possess the same outer radius; hence, an interesting research topic will be to utilize additional information such as colour and texture to further refine the classification based on material type or usage.
3- The use of autonomous robots on construction sites to record videos and point clouds with the aid of a planned BIM for initial trajectory planning is an interesting and cutting-edge topic.
4- New developments in smartphone technologies, such as laser scanning on the iPhone 12 and 8K video recording using the Samsung S20 series, offer additional possibilities for future research. For instance, the accuracy of the point cloud generated using the combination of laser scanner and the video recordings is an interesting subject for investigation. Furthermore, the effectiveness of the scale definition through the built-in laser scanner of iPhone 12 compared to our proposed process for scale definition can also be examined.

**Acknowledgements**

The authors wish to acknowledge the support provided by the MJS Mechanical Ltd. and Michael Baytalan for their cooperation and professional installation of the mechanical mock pipes and construction site access for the purpose of the experiments, presented in this manuscript, along with the methodological improvements proposed by thorough discussions with the members of the DICE-Technologies team. This research project was partly funded by the Natural Sciences and Engineering Research Council (NSERC) of Canada (542980 - 19) and Alberta Innovates (G2020000051).